\documentclass{article}
\usepackage{spconf,amsmath,graphicx,amsfonts,xcolor,soul}
\sethlcolor{yellow}

\title{A novel multimodal dynamic fusion network for disfluency detection in spoken utterances}
\name{Sreyan Ghosh$^{1\star}$, Utkarsh Tyagi$^{2\star}$, Sonal Kumar$^{3\star}$, Manan Suri$^{2,4\star}$, Rajiv Ratn Shah$^2$\thanks{\hspace*{-1mm}$^{\star}$These authors contributed equally to this work}}

\address{
  $^1$University of Maryland, College Park, USA, $^2$MIDAS Labs, IIIT-Delhi, \\
  $^3$Cisco Systems, Bangalore, $^4$NSUT Delhi, India}
\begin{document}
%
\maketitle
\begin{abstract}

\end{abstract}

Disfluency, though originating from human spoken utterances, is primarily studied as a uni-modal text-based Natural Language Processing (NLP) task. Based on early-fusion and self-attention-based multimodal interaction between text and acoustic modalities, in this paper, we propose a novel multimodal architecture for disfluency detection from individual utterances. Our architecture leverages a multimodal dynamic fusion network that adds minimal parameters over an existing text encoder commonly used in prior art to leverage the prosodic and acoustic cues hidden in speech. Through experiments, we show that our proposed model achieves state-of-the-art results on the
widely used English Switchboard for disfluency detection and outperforms prior unimodal and multimodal systems in literature by a significant margin. In addition, we make a thorough qualitative analysis and show that, unlike text-only systems, which suffer from spurious correlations in the data, our system overcomes this problem through additional cues from speech signals. We make all our codes publicly available on GitHub \footnote{We will release code on paper acceptance}.

\section{Introduction}
\label{sec:intro}

Unlike written text, in spoken conversations, humans often fail to pre-plan what they are about to speak. This phenomenon, known as disfluency, is a para-linguistic concept that is ubiquitous in human conversations. In the past decade, with the rapid adoption of speech as an input to modern intelligent Natural Language Understanding (NLU) systems, disfluency detection has become increasingly popular as an emerging research topic as NLU systems trained on fluent data can easily get misled due to the presence of disfluencies.

\begin{figure}[t]
  \centering
  \includegraphics[height=30pt, width=180pt]{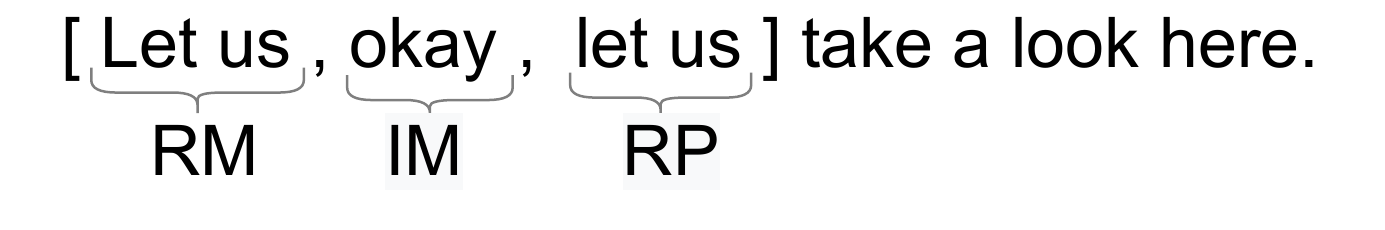}
  \caption{\textit{A sentence from the English Switchboard
corpus with disfluencies. RM=Reparandum,
IM=Interregnum, RP=Repitition. The preceding RM is corrected by the following RP.
}}
  \label{fig:dis_fig}
\end{figure}

Fig. \ref{fig:dis_fig} shows an example of disfluency, whereby it can be divided into 3 main parts, a reparandum, an optional interregnum, and a repair. Disfluency detection as a Natural Language Processing (NLP) task generally focuses on identifying and removing reparandum. Furthermore, disfluencies can primarily be categorized into 5 types, namely, repetitions, restarts, repairs, deletions, and substitutions (see Table \ref{tab:dis_table} for examples). Repetition primarily occurs when linguistic components repeat, usually in the form of partial words, complete words, or short phrases. Substitution occurs when the same linguistic components are replaced in order to clarify an underlying concept. Deletions or false restart refers to abandoned linguistics components.

\begin{table}[t]
    \centering
    \caption{Different types of disfluencies.}
    \begin{tabular}{|l|l|}
        \hline Type & Example \\
        \hline Repair & [i do + i] ski yes \\
        \hline Repetition & but [i + i] grew up with cats \\
        \hline Restart & [you were +$\{$uh$\}$ ] he was waiting \\ & for what again \\
        \hline Deletion & [i that it just +] you know it 's \\ & absolutely  devastating \\
        \hline Substitution & the pen was kept [under + over] \\ & the table\\
        \hline
    \end{tabular}
    \label{tab:dis_table}
\end{table}

In the past decade, neural architectures have shown promising results in the task of disfluency detection, where the majority of prior work primarily reports results on the SwitchBoard (SWBD) corpus \cite{godfrey1992switchboard}. The Switchboard corpus consists of telephone conversations between speakers of American English and is transcribed for fully spontaneous speech, which makes it suitable for this task. However, a majority of these systems achieving state-of-the-art performance, leverage the modality of text only to detect reparandums solving a sequence tagging task in segmented transcripts from the SWBD corpus. Though the modality of text has rich semantic and syntactic information and contextualized token representations from transformer models aid systems in recognizing out-of-linguistic-context words like disfluencies, we acknowledge the fact that disfluency generally originates from spoken utterances and acoustic cues like lexical, prosody, pitch, stutter, etc. which are largely ignored can prove as important signals for disfluency detection. Though some amount of prior work for this task leverage the modality of speech \cite{zayats2019giving,ferguson-etal-2015-disfluency,deng-etal-2020-integrating,tanaka2019disfluency}, most of these systems suffer from 2 main problems: 1) They require hand-engineering of certain acoustic and prosodic features and the performance depends heavily on the quality of features mined. 2) They rely on a simple concatenation of text and acoustic features and fail to pay attention to fine-grained multimodal information, thus making them unable to capture inter-modality interactions effectively.

{\noindent \textbf{Main Contribution:}} To address these gaps, we propose a novel multimodal framework MDFN that leverages the modalities of both speech and text as input without requiring the need for specific hand-engineered features. To achieve this, we first adopt the use of pre-trained contextualized representation models for both the text and speech modalities separately, where we make use of BERT \cite{devlin2018bert} architecture as the text encoder and wav2vec-2.0 \cite{baevski2020wav2vec} as the speech encoder. Second, to better capture the implicit alignments between individual text tokens and speech frames, we propose the use of a unique 2 branch multimodal interaction module (MMI). MMI essentially couples the standard Transformer layer with a cross-modal attention mechanism to produce a speech-aware word representation and a word-aware speech representation for each input word. Next, we concatenate the utterance representation obtained from both branches to finally classify disfluent tokens in a span classification setup. Our system achieves state-of-the-art results for disfluency on the SWBD dataset. Additionally, we also make a thorough qualitative analysis to show the effectiveness of speech signals in disfluency detection.


\section{Related Work}
\label{sec:relatedwork}


Four major categories can be used to categorize disfluency detection models. The first one uses noisy channel models \cite{zwarts-johnson-2011-impact,jamshid-lou-etal-2018-disfluency}, which call for a transducer in the channel model based on Tree Adjoining Grammar (TAG). The second group makes use of phrase structure, which is frequently connected to transition-based parsing but calls for annotated syntactic structure~\cite{lou2020improving,rasooli-tetreault-2013-joint}. The last category uses end-to-end Encoder-Decoder models to automatically recognize disfluent segments, while the third and most popular category frames the work as a sequence tagging problem \cite{ferguson2015disfluency,hough2015recurrent}. In recent work in this area, the authors of \cite{yang-etal-2020-planning,wang2020combining} offer a self-supervised learning and data augmentation technique to learning disfluencies that have been demonstrated to decrease the gap with supervised training, reducing the dependence of disfluency detection on human-annotated datasets.

Learning disfluency detection is often framed as a sequence tagging task \cite{yang-etal-2020-planning,wang2020combining,ferguson2015disfluency,ghosh22c_interspeech,lou2020improving,lee21g_interspeech}. As evident from prior work in this domain, most disfluency models leverage only text transcripts from spoken utterances, with some exceptions being \cite{zayats2019giving,ferguson-etal-2015-disfluency,deng-etal-2020-integrating,tanaka2019disfluency} where the authors use acoustic and prosodic features from speech together with token-level features from text to predict disfluent words. Beyond reporting improvements in performance over just using text, the authors also show how using prosodic features improves the detection of complex, long, and fake disfluencies. However, these features are hand-crafted. In recent times, Self-supervised learning (SSL) for speech has proven to be very effective on a wide range of Spoken Language Processing (SLP) tasks \cite{yang2021superb}. Inspired by these findings, we propose to use high-level speech representations from SSL models for disfluency detection.



\begin{figure}[t]
  \centering
\includegraphics[width=7.5cm,height=11cm,trim={0.5cm 0cm 0cm 0cm}]{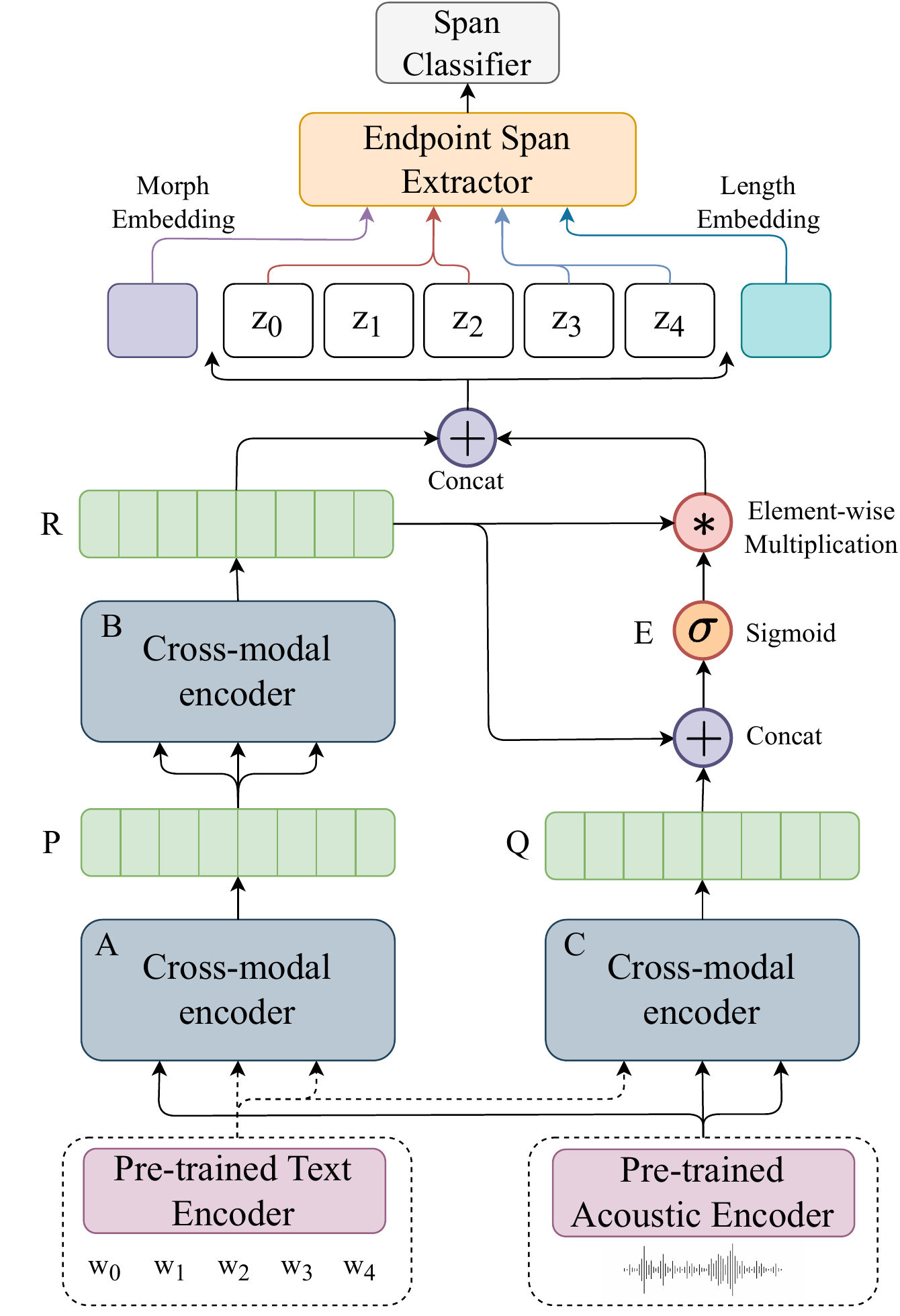}
  \caption{Illustration of the MDFN architecture.}\label{fig:figure_2}
\end{figure}

\section{Methodology}
\label{sec:methodology}

\subsection{Problem Formulation}
Suppose we have a dataset $D$ with $N$ utterances \{$u$\textsubscript{$1$}, $u$\textsubscript{$2$}, $u$\textsubscript{$3$}, $\cdots$, $u$\textsubscript{$N$}\} where each utterance $u_i$ has both speech cues $a_i$ and text cues $t_i$ available where $u_i$ $=$ ($a_i$,$t_i$). $t_i$ can be ASR transcripts or human-annotated transcripts. We denote the \emph{i}-th sentence with \emph{M} tokens as \emph{t\textsubscript{i}} = \{$w$\textsubscript{$t$}$|$ $t$ = 1, $\cdots$ $T$\}. The corresponding label set is defined by \{$d\textsubscript{1}$,$d\textsubscript{2}$,$d\textsubscript{3}$, $\cdots$ $d$\textsubscript{$N$}\} where $d\textsubscript{i}$ is the label sequence for each sentence and is denoted by $d\textsubscript{i}$ = \{$y$\textsubscript{$t$}$|$ $t$ = 1, $\cdots$ $M$\} where $y$\textsubscript{$t$} $\in \mathcal{Y}$ and $\mathcal{Y}$ $=$ $\{I,O\}$. $I$ here stands for disfluent tokens and $O$ stands for fluent tokens.

\subsection{Contextualized Representations}
\label{sec:context}


{\textbf{Text Representations:}} Following much of prior art \cite{ghosh22c_interspeech,lee21g_interspeech}, we use  BERT\textsubscript{\emph{BASE}} \cite{devlin2019bert} from the transformers family as our contextualized text encoder to encode the transcripts for human-annotated utterances and obtain rich contextualized token representations. Thus for a total of $M$ tokens, the $m\textsuperscript{th}$ contextualised embedding for each token in text transcript $t_i$, of utterance $u_i$, is denoted by $e_{m}^{t_{i}}$ $\in$ $\mathbb{R}^{768}$.

\vspace{1.5mm}
{\noindent \textbf{Speech Representations:}} We obtain context-aware powerful speech representations from wav2vec 2.0 \cite{baevski2020wav2vec}. Wav2vec 2.0 is pre-trained using SSL on unlabeled raw audio. For more details on pre-training and architecture of wav2vec 2.0, we refer our readers to \cite{baevski2020wav2vec}. Wav2vec 2.0 outputs $J$ hidden states and we denote the $j\textsuperscript{th}$ hidden state or contextualized embeddings from the raw audio input $a_i$ of utterance $u_i$ as $e_{j}^{a_{i}}$ where $e_{j}^{a_{i}}$ $\in$ $\mathbb{R}^{768}$ . $J$ depends on the length of the raw audio file and the CNN feature extraction layer of wav2vec 2.0, which extracts frames from the raw audio with a stride of 20ms and a hop size of 25ms.

\subsection{Multimodal Interaction Module (MMI)}
\label{sec:multimodal-prediction}

Our Multimodal Interaction Module (MMI) consists of 3 Cross-Modal Encoder (CME) blocks annotated as $A$, $B$, and $C$ in Fig. \ref{fig:figure_2}. Each of these 3 CME blocks is similar to a generic transformer layer \cite{vaswani2017attention}, where each layer is composed of an \emph{h}-head $\mathbf{CMA}$ module \cite{tsai2019multimodal}, residual connections, and feed-forward layers.

\vspace{1.5mm}
{\noindent \textbf{Speech-Aware Word Representations:}} As shown in Fig.\ref{fig:figure_2}, to learn token representations that are aware or knowledgeable of the associated spoken utterance, we feed wav2vec-2.0 embeddings $\mathbf{A}$ $\in$ $\mathbb{R}^{d \times J}$ as queries and token embeddings $\mathbf{T}$ $\in$ $\mathbb{R}^{d \times M}$ as keys and values into $\mathbf{CMA}$ module of $\mathbf{CME}$ block $A$ as follows:

\begin{equation}
    \mathbf{CMA}(\mathbf{A}, \mathbf{T})=\operatorname{softmax}\left(\frac{\left[\mathbf{W}_{\mathbf{q}_{\mathbf{i}}} \mathbf{A}\right]^{\top}\left[\mathbf{W}_{\mathbf{k}_{\mathbf{i}}} \mathbf{T}\right]}{\sqrt{d / m}}\right)\left[\mathbf{W}_{\mathbf{v}_{\mathbf{i}}} \mathbf{T}\right]^{\top}
\end{equation}

where \{$\mathbf{W}_{\mathbf{q}_{\mathbf{i}}}$, $\mathbf{W}_{\mathbf{k}_{\mathbf{i}}}$, $\mathbf{W}_{\mathbf{v}_{\mathbf{i}}}$\} $\in$ $\mathbb{R}^{d/m \times h}$ denote the query, key and value weight matrices respectively for the $i^{th}$ attention head. The final output representation of the $\mathbf{CME}$ block $B$ is now $\mathbf{P}$ $=$ ($\mathbf{p_0}$, $\mathbf{p_1}$, $\cdots$, $\mathbf{p_{m-1}}$). Post this step, to address the fact that each generated representation $\mathbf{p}_i$ in the previous block corresponds to the $i^{th}$ acoustic embedding and not the token embedding; we feed $\mathbf{P}$ to another $\mathbf{CME}$ block $B$, which treats the original token embeddings $\mathbf{T}$ as queries and $\mathbf{P}$ as keys and values. Finally, we now obtain the final Speech-Aware Word Representations as $\mathbf{R}$ $=$ ($\mathbf{r_0}$,$\mathbf{r_1}$, $\cdots$, $\mathbf{r_{j-1}}$).

\vspace{1.5mm}
{\noindent \textbf{Word-Aware Speech Representations:}} Next, to obtain speech representations that are aware of their corresponding semantic token representations and align each word to its closely related wav2vec-2.0 embeddings, we make use of another $\mathbf{CME}$ block $D$ by treating $\mathbf{T}$ as queries and $\mathbf{A}$ as keys and values. The final representations obtained from the block can be denoted as $\mathbf{Q}$ $=$ ($\mathbf{q_0}$, $\mathbf{q_1}$, $\cdots$, $\mathbf{q_{j-1}}$).

\vspace{1.5mm}
{\noindent \textbf{Acoustic Gate:}} In practice, speech frames might encode redundant information like random noise, and this makes it an important step to implement an acoustic gate $E$ which can dynamically control the contribution of each speech frame embedding. Following much work in literature, we implement an acoustic gate $\mathbf{g}$ as follows:

\begin{equation}
\mathbf{g}=\sigma\left(\mathbf{W}_{g}^{\top}[ \mathbf{R} ; \mathbf{Q}] + \mathbf{B}_{g}\right)
\end{equation}

where $\sigma$ is the element-wise sigmoid function. Finally, based on the gate output, the final word-aware speech representations are obtained by $\mathbf{Q}$ = $\mathbf{g} . \mathbf{Q}$.

\subsection{Span Classification}

The final span representations from both branches of the MMI module are then concatenated to obtain our final cross-modal MMI representations $\mathbf{M}$ $\in$ $\mathbb{R}^{2d}$ where $\mathbf{M}$ = [$\mathbf{Q}$ ; $\mathbf{R}$]. Inspired by \cite{ghosh22c_interspeech}, for our system, we choose to solve a span-level classification task over token-level classification. Thus after we obtain $\mathbf{M}$, we enumerate through all possible $k$ spans $J$ = \{$j$\textsubscript{1}, $\cdots$ , $j$\textsubscript{i}, $\cdots$ , $j$\textsubscript{$k$}\} up to a maximum number of consecutive tokens and then re-assign a
label $y_i$ $\in$ $\{I,O\}$ for each span $j_i$. We then formulate the vectorial representation of each span as the concatenation of the representations of the starting token $\mathbf{m}_{b_i}$ $\in \mathbb{R}^{2d}$, the ending token $\mathbf{m}_{s_i}$ $\in \mathbb{R}^{2d}$, and a length embedding $\mathbf{\ell}_{i}$ $\in \mathbb{R}^{len}$.The length embedding is implemented as a look-up table similar to \cite{ghosh22c_interspeech}. The final vector representation for each span fed into the span prediction layer is now $\mathbf{j}_i$ $=$ [$\mathbf{m}_{b_i}$ $;$ $\mathbf{m}_{s_i}$ $;$ $\mathbf{\ell}_{i}$]. $\mathbf{j}_i$ is then passed through a linear transformation followed by a $\operatorname{softmax}$ operation to find the final class among $\{I,O\}$ for each utterance. Post this step, we employ the heuristic decoding method proposed by \cite{ghosh22c_interspeech}.




\section{Experiments}
\label{sec:experiments}

\subsection{Dataset}
\label{sec:dataset}
We evaluate our proposed system on the human-annotated disfluency annotations \cite{zayats2019disfluencies} for the English Switchboard Dataset \cite{godfrey1992switchboard}. For fair evaluation, following much of prior-art \cite{charniak2001edit}, we split the entire SWBD dataset into training (sw23$[\star]$.dps), development (sw4[5-9]$[\star]$.dps), and test (sw4[0-1]$[\star]$.dps) splits. Post this, following \cite{hough2015recurrent}, we pre-process the text and convert all all characters to lowercase and remove all punctuation and partial words.

\subsection{Experimental Setup}
\label{sec:experimental_setup}
All our models are implemented using the PyTorch deep learning framework. We use the BERT\textsubscript{BASE} model as our pre-trained contextualized token embedding model and the robust wav2vec 2.0\textsubscript{LARGE} \cite{hsu2021robust} fine-tuned on SWBD as our contextualized acoustic embedding model. Pre-trained checkpoints and implementations for both models are adopted from the Huggingface library. We train our MDFN model for 20 epochs with a batch size of 32 using adam optimizer with a learning rate of $1 \times 10^{-5}$. The dimension of our span length embedding $\mathbb{R}^{len}$ is 100, and the morph embedding is 100.

\subsection{Baselines and Compared Methods}
For comparison to prior art, we resort to both unimodal ad multimodal sequence tagging baselines evaluated under the \emph{IO} tagging scheme. For our text-only unimodal baseline, we resort to a simple span classifier BERT system from the transformers family also evaluated on the \emph{IO} tagging scheme.

\section{Results}
\label{sec:results}
Table \ref{tab:result_table} shows the evaluation results on the SWBD test set compared to the prior art on disfluency detection. Following much of prior art, we evaluate the performance of our proposed MDFN model on \emph{$F_{1}$}. As we clearly see, the our proposed model outperforms the current state-of-the-art (SOTA) system \cite{ghosh22c_interspeech} in $F_1$ scores by 1.5\%. With a slightly higher recall, our model is more sensitive toward the positive class but nevertheless maintains high precision and hence a high $F_1$ score.  Additionally, in the next section, we also make an effort to analyze the benefits of using multimodal interaction information in our span classifier model.

\begin{table}[ht]
    \centering
    \caption{Evaluation results of our proposed model compared to the baselines and prior art on the Switchboard test set. The best scores are denoted in bold.}
        \vspace{1mm}
        \begin{tabular}{l|c|c|c}
        \hline \textbf{Model} & $\mathrm{P}$ & $\mathrm{R}$ & $\mathrm{F\textsubscript{1}}$\\
        \hline Semi-CRF \cite{ferguson2015disfluency} & $90.0$ & $81.2$ & $85.4$ \\
        Bi-LSTM \cite{zayats2016disfluency} & $91.6$ & $80.3$ & $85.9$ \\
        Attention-based \cite{wang-etal-2016-neural} & $91.6$ & $82.3$ & $86.7$ \\
        Transition-based \cite{wang2017transition} & $91.1$ & $84.1$ & $87.5$ \\
        Self-supervised \cite{wang2020combining} & $93.4$ & $87.3$ & $90.2$ \\
        Self-trained \cite{lou2020improving} & $87.5$ & $93.8$ & $90.6$ \\
        EGBC \cite{bach2019noisy} & $95.7$ & $88.3$ & $91.8$ \\
        BERT fine-tune \cite{bach2019noisy} & $94.7$ & $89.8$ & $92.2$ \\
        BERT-CRF-Aux \cite{lee21g_interspeech} & $94.6$ & $91.2$ & $92.9$ \\
        ELECTRA-CRF-Aux \cite{lee21g_interspeech} & $94.8$ & $91.6$ & $93.1$\\
        Span Classification BERT-GCN\cite{ghosh22c_interspeech}  & $\mathbf{95.2}$ & $93.2$ & $94.2$ \\
        BERT (Baseline) & $95.1$ & $93$ & $94.1$ \\
        \textbf{MDFN} (ours) & $92.8$ & $\mathbf{98.7}$ & $\mathbf{95.7}$ \\
        \hline
        \end{tabular}
    \label{tab:result_table}
\end{table}

\section{Result Analysis}
\label{sec:result_analysis}
\textbf{MDFN vs. Span Classification BERT-GCN:} Highlighted words below represent the predictions made by the state-of-the-art text-only \emph{Span Classification BERT-GCN} (SCBG) model \cite{ghosh22c_interspeech} from literature. In contrast, our MDFN model predicted that the sentences are fluent, which is in line with the ground truth annotations.
\begin{enumerate}
    \item i finally got impaneled on one case \hl{on my} next to the last day
    \item and \hl{that is} that money tends to stick where it lands first
\end{enumerate}
As we see in both examples, the spans classified as disfluent by the SCBG are not disfluencies according to ground truth and our model. We hypothesize that although \cite{ghosh22c_interspeech} captures and repairs the longer restarts, it wrongly classifies spoken utterances which contain nuances like incorrect grammar and usage of multiple conjunctions, which makes the structure of the sentence complex and seemingly disfluent. For example ``on'' and ``that'' appears twice in immediate succession in sentences 1 and 2, respectively. Thus we can conclude that text-only \cite{ghosh22c_interspeech} gets biased to such spurious correlations, which makes it identify these words as repetitions. In such cases, the acoustic modality helps capture the true nature of the sentence with the help of the confidence and tone of the speaker in audio.

Highlighted words below represent the predictions made by our MDFN model. In contrast, the SCBG model predicted the sentences as fluent, which is in line with the ground truth annotations.
\begin{enumerate}
    \item \hl{end} it 's very easy
    \item \hl{any} so how do you get most of your news
\end{enumerate}
In the switchboard dataset, there are multiple instances of segmented sentences which occur due to cut-offs or speaker overlap. Due to this, the initial or end utterances can be wrongly classified as disfluent, as the MDFN model doesn't use linguistic rules to find the relation among words. These are captured by parse trees used by the SCBG model, as seen in the above examples.

\section{Conclusion}
\label{sec:conclusion}

In this paper, we present a novel multimodal dynamic fusion network that effectively utilizes both the modalities of speech and text for disfluency detection. Our proposed system achieves state-of-the-art results in disfluency detection on the very popular SWBD dataset. As part of future work, we would like to propose better multimodal networks for disfluency detection that better capture the implicit relations between the two modalities. Additionally, we would also like to perform a human evaluation of our model results for ASR-generated transcripts to evaluate the effectiveness of our model over unimodal text approaches in such a setting.

\newpage
\bibliographystyle{IEEEbib}
\bibliography{strings,refs}

\end{document}